\documentclass[conference]{IEEEtran}
\usepackage{times}

\newcommand\nnfootnote[1]{%
  \begin{NoHyper}
  \renewcommand\thefootnote{}\footnote{#1}%
  \addtocounter{footnote}{-1}%
  \end{NoHyper}
}

\usepackage[numbers]{natbib}
\usepackage{multicol}
\usepackage[bookmarks=true]{hyperref}

\usepackage{amsmath} 
\usepackage{amssymb}  

\usepackage{verbatim}

\usepackage{mathtools}
\usepackage[ruled,linesnumbered]{algorithm2e}

\usepackage{amssymb}

\usepackage{rotating}
\usepackage{graphicx}
\usepackage{caption}
\usepackage{subcaption}
\usepackage{comment}
\usepackage[colorinlistoftodos]{todonotes}

\usepackage{fancybox,color}

\definecolor{grey}{rgb}{0.5,0.5,0.5}

\usepackage[utf8]{inputenc}
\usepackage[english]{babel}

\begin{document}

\title{EV-FlowNet: Self-Supervised Optical Flow Estimation for Event-based Cameras}



%
\author{\authorblockN{Alex Zihao Zhu,
Liangzhe Yuan,
Kenneth Chaney and
Kostas Daniilidis}
\authorblockA{School of Engineering and Applied Science\\University of Pennsylvania\\\{alexzhu, lzyuan, chaneyk, kostas\}@seas.upenn.edu}}
\maketitle
\begin{abstract}
Event-based cameras have shown great promise in a variety of situations where frame based cameras suffer, such as high speed motions and high dynamic range scenes. However, developing algorithms for event measurements requires a new class of hand crafted algorithms. Deep learning has shown great success in providing model free solutions to many problems in the vision community, but existing networks have been developed with frame based images in mind, and there does not exist the wealth of labeled data for events as there does for images for supervised training. To these points, we present EV-FlowNet, a novel self-supervised deep learning pipeline for optical flow estimation for event based cameras. In particular, we introduce an image based representation of a given event stream, which is fed into a self-supervised neural network as the sole input. The corresponding grayscale images captured from the same camera at the same time as the events are then used as a supervisory signal to provide a loss function at training time, given the estimated flow from the network. We show that the resulting network is able to accurately predict optical flow from events only in a variety of different scenes, with performance competitive to image based networks. This method not only allows for accurate estimation of dense optical flow, but also provides a framework for the transfer of other self-supervised methods to the event-based domain.
\end{abstract}
\IEEEpeerreviewmaketitle
\nnfootnote{Associated dataset: \url{https://daniilidis-group.github.io/mvsec/}.}
\nnfootnote{Supplementary video: \url{https://youtu.be/eMHZBSoq0sE}.}
\section{Introduction}
\label{sec:introduction}

By registering changes in log intensity in the image with microsecond accuracy, event-based cameras offer promising advantages over frame based cameras in situations with factors such as high speed motions and difficult lighting. One interesting application of these cameras is the estimation of optical flow. By directly measuring the precise time at which each pixel changes, the event stream directly encodes fine grain motion information, which researchers have taken advantage of in order to perform optical flow estimation. For example, \citet{benosman2012asynchronous} show that optical flow can be estimated from a local window around each event in a linear fashion, by estimating a plane in the spatio-temporal domain. This is significantly simpler than image-based methods, where optical flow is performed using iterative methods. However, analysis in \citet{rueckauer2016evaluation} has shown that these algorithms require significant, hand crafted outlier rejection schemes, as they do not properly model the output of the sensor.

For traditional image-based methods, deep learning has helped the computer vision community achieve new levels of performance while avoiding having to explicitly model the entire problem. However, these techniques have yet to see the same level of adoption and success for event-based cameras. One reason for this is the asynchronous output of the event-based camera, which does not easily fit into the synchronous, frame-based inputs expected by image-based paradigms. Another reason is the lack of labeled training data necessary for supervised training methods. In this work, we propose two main contributions to resolve these issues. 

\begin{figure}[t!]
\centering
  \includegraphics[trim=0 0 0 -1cm, width=0.48\linewidth]{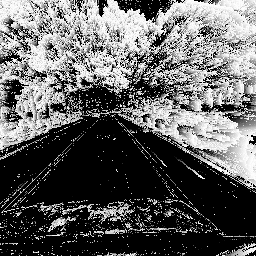}
  \includegraphics[trim=0 0 0 -1cm, width=0.48\linewidth]{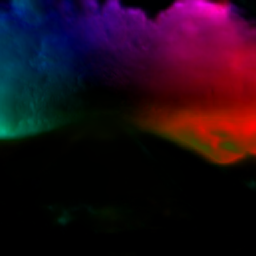}
  \caption{Left: Event input to the network visualizing the last two channels (latest timestamps). Right: Predicted flow, colored by direction. Best viewed in color.}
  \label{fig:timestamp_image}
\end{figure}

First, we propose a novel image-based representation of an event stream, which fits into any standard image-based neural network architecture. The event stream is summarized by an image with channels representing the number of events and the latest timestamp at each polarity at each pixel. This compact representation preserves the spatial relationships between events, while maintaining the most recent temporal information at each pixel and providing a fixed number of channels for any event stream.

Second, we present a self-supervised learning method for optical flow estimation given only a set of events and the corresponding grayscale images generated from the same camera. The self-supervised loss is modeled after frame based self-supervised flow networks such as \citet{jason2016back} and \citet{meister2017unflow}, where a photometric loss is used as a supervisory signal in place of direct supervision. As a result, the network can be trained using only data captured directly from an event camera that also generates frame based images, such as the Dynamic and Active-pixel Vision (DAVIS) Sensor developed by \citet{brandli2014240}, circumventing the need for expensive labeling of data.

These event images combined with the self-supervised loss are sufficient for the network to learn to predict accurate optical flow from events alone. For evaluation, we generate a new event camera optical flow dataset, using the ground truth depths and poses in the Multi Vehicle Event Camera Dataset by \citet{zhu2018mvsec}. We show that our method is competitive on this dataset with UnFlow by \citet{meister2017unflow}, an image-based self supervised network trained on KITTI, and fine tuned on event camera frames, as well as standard non-learning based optical flow methods.

In summary, our main contributions in this work are:
\begin{itemize}
\item
We introduce a novel method for learning optical flow using events as inputs only, without any supervision from ground-truth flow.
\item
Our CNN architecture uses a self-supervised photoconsistency loss from low resolution intensity images used in training only.
\item
We present a novel event-based optical flow dataset with
ground truth optical flow, on which we evaluate our
method against a state of the art frame based method.
\end{itemize}
\section{Related Work}
\label{sec:relatedwork}
\subsection{Event-based Optical Flow}
There have been several works that attempt to take advantage of the high temporal resolution of the event camera to estimate accurate optical flow. 
\citet{benosman2012asynchronous} model a given patch moving in the spatial temporal domain as a plane, and estimate optical flow as the slope of this plane. This work is extended in \citet{benosman2014event} by adding an iterative outlier rejection scheme to remove events significantly far from the plane, and in \citet{barranco2014contour} by combining the estimated flow with flow from traditional images. \citet{brosch2015event} present an analogy of \citet{lucas1981iterative} using the events to approximate the spatial image gradient, while \citet{orchard2014bioinspired} use a spiking neural network to estimate flow, and \citet{liu2018abmof} estimate sparse flow using an adaptive block matching algorithm. In other works, \citet{bardow2016simultaneous} present the optical flow estimation problem jointly with image reconstruction, and solve the joint problem using convex optimization methods, while \citet{zhu2017event} present an expectation-maximization based approach to estimate flow in a local patch. A number of these methods have been evaluated in \citet{rueckauer2016evaluation} against relatively simple scenes with limited translation and rotation, with limited results, with ground truth optical flow estimated from a gyroscope. Similarly, \citet{barranco2016dataset} provide a dataset with optical flow generated from a known motion combined with depths from a RGB-D sensor.	

\subsection{Event-based Deep Learning}
One of the main challenges for supervised learning for events is the lack of labeled data. As a result, many of the early works on learning with event-based data, such as 
\citet{ghosh2014real} and \citet{moeys2016steering}, rely on small, hand collected datasets. 

To address this, recent works have attempted to collect new datasets of event camera data. \citet{mueggler2017event}, provide handheld sequences with ground truth camera pose, which \citet{nguyen2017real} use to train a LSTM network to predict camera pose. In addition, \citet{zhu2018mvsec} provide flying, driving and handheld sequences with ground truth camera pose and depth maps, and \citet{DBLP:journals/corr/abs-1711-01458} provide long driving sequences with ground truth measurements from the vehicle such as steering angle and GPS position.

Another approach has been to generate event based equivalents of existing image based datasets by recording images from these datasets from an event based camera (\citet{orchard2015converting}, \citet{hu2016dvs}). 

Recently, there have also been implementations of neural networks on spiking neuromorphic processors, such as in \citet{amir2017low}, where a network is adapted to the TrueNorth chip to perform gesture recognition.

\subsection{Self-supervised Optical Flow}
Self-supervised, or unsupervised, methods have shown great promise in training networks to solve many challenging 3D perception problems. \citet{jason2016back} and \citet{ren2017unsupervised} train an optical flow prediction network using the traditional brightness constancy and smoothness constraints developed in optimization based methods such as the Lucas Kanade method \citet{lucas1981iterative}. \citet{zhu2017guided} combine this self-supervised loss with supervision from an optimization based flow estimate as a proxy for ground truth supervision, while \citet{meister2017unflow} extend the loss with occlusion masks and a second order smoothness term, and \citet{lai2017semi} introduce an adversarial loss on top of the photometric error. 
\begin{figure}[t!]
\centering
  \includegraphics[trim=0 0 0 -1cm, width=0.48\linewidth]{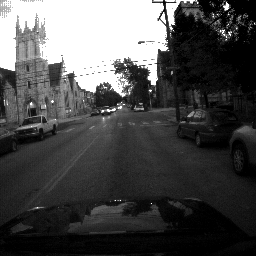}
  \includegraphics[trim=0 0 0 -1cm, width=0.48\linewidth]{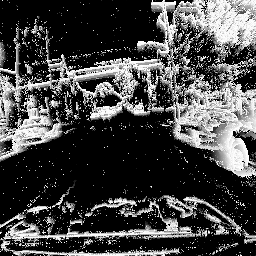}
  \caption{Example of a timestamp image. Left: Grayscale output. Right: Timestamp image, where each pixel represents the timestamp of the most recent event. Brighter is more recent.}
  \label{fig:timestamp_image}
\end{figure}
\begin{figure*}[t!]
\centering
  \includegraphics[trim=0 0 0 -1cm, width=0.8\linewidth]{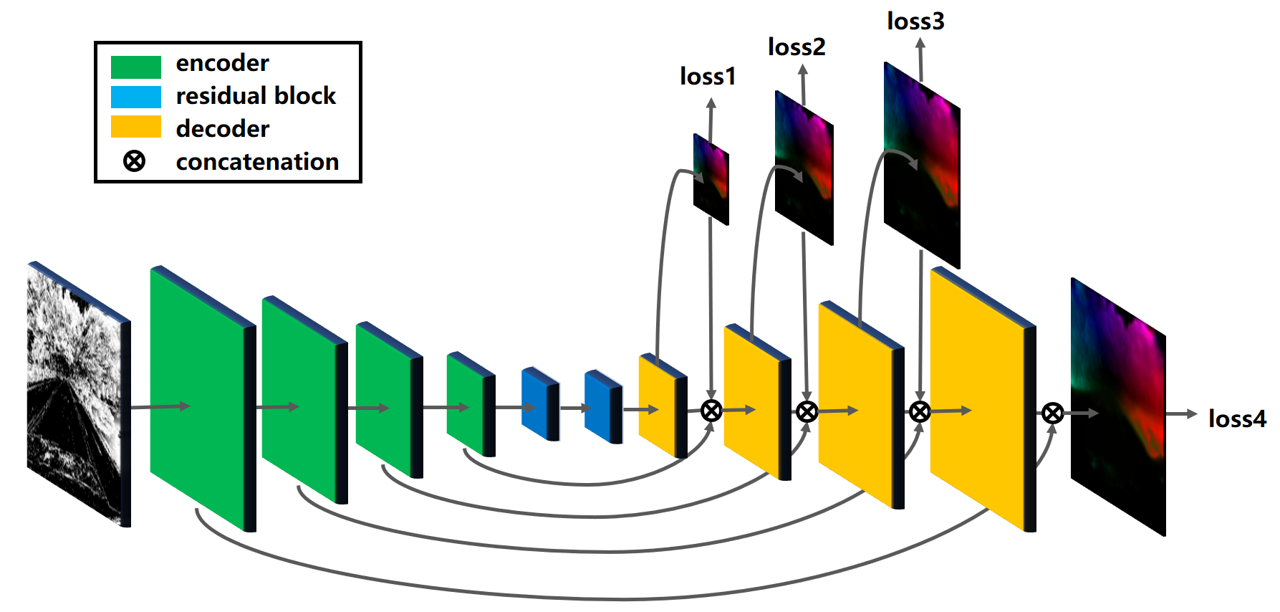}
  \caption{EV-FlowNet architecture. The event input is downsampled through four encoder (strided convolution) layers, before being passed through two residual block layers. The activations are then passed through four decoder (upsample convolution) layers, with skip connections to the corresponding encoder layer. In addition, each set of decoder activations is passed through another depthwise convolution layer to generate a flow prediction at its resolution. A loss is applied to this flow prediction, and the prediction is also concatenated to the decoder activations. Best viewed in color.}
  \label{fig:architecture}
\end{figure*}
\section{Method}
\label{sec:method}
In this section, we describe our approach in detail. In Sec. \ref{sec:representation}, we describe our event representation, which is an analogy to an event image. In Sec. \ref{sec:loss}, we describe the self-supervised loss used to provide a supervisory signal using only the gray scale images captured before and after each time window, and in Sec. \ref{sec:architecture}, we describe the architecture of our network, which takes as input the event image and outputs a pixel-wise optical flow. Note that, throughout this paper, we refer to optical flow as the displacement of each pixel within a given time window. 
\subsection{Event Representation}
\label{sec:representation}
An event-based camera tracks changes in the log intensity of an image, and returns an event whenever the log intensity changes over a set threshold $\theta$:
\begin{align}
\log(I_{t+1})&-\log(I_t) \geq \theta
\intertext{Each event contains the pixel location of the change, timestamp of the event and polarity:}
e=&\begin{Bmatrix}\mathbf{x},&t,&p\end{Bmatrix}
\end{align}
Because of the asynchronous nature of the events, it is not immediately clear what representation of the events should be used in the standard convolutional neural network architecture. Most modern network architectures expect image-like inputs, with a fixed, relatively low, number of channels (recurrent networks excluded) and spatial correlations between neighboring pixels. Therefore, a good representation is key to fully take advantage of existing networks while summarizing the necessary information from the event stream.

Perhaps the most complete representation that preserves all of the information in each event would be to represent the events as a $n\times 4$ matrix, where each column contains the information of a single event. However, this does not directly encode the spatial relationships between events that is typically exploited by convolutions over images. 

In this work, we chose to instead use a representation of the events in image form. The input to the network is a 4 channel image with the same resolution as the camera. 

The first two channels encode the number of positive and negative events that have occurred at each pixel, respectively. This counting of events is a common method for visualizing the event stream, and has been shown in \citet{nguyen2017real} to be informative in a learning based framework to regress 6dof pose.

However, the number of events alone discards valuable information in the timestamps that encode information about the motion in the image. Incorporating timestamps in image form is a challenging task. One possible solution would be to have $k$ channels, where $k$ is the most events in any pixel in the image, and stack all incoming timestamps. However, this would result in a large increase in the dimensionality of the input. Instead, we encode the pixels in the last two channels as the timestamp of the most recent positive and negative event at that pixel, respectively. This is similar to the "Event-based Time Surfaces" used in \citet{lagorce2017hots} and the "timestamp images" used in \citet{park2016performance}. An example of this kind of image can be found in Fig.~\ref{fig:timestamp_image}, where we can see that the flow is evident by following the gradient in the image, particularly for closer (faster moving) objects. While this representation inherently discards all of the timestamps but the most recent at each pixel, we have observed that this representation is sufficient for the network to estimate the correct flow in most regions. One deficiency of this representation is that areas with very dense events and large motion will have all pixels overridden by very recent events with very similar timestamps. However, this problem can be avoided by choosing smaller time windows, thereby reducing the magnitude of the motion.

In addition, we normalize the timestamp images by the size of the time window for the image, so that the maximum value in the last two channels is 1. This has the effect of both scaling the timestamps to be on the same order of magnitude as the event counts, and ensuring that fast motions with a small time window and slow motions with a large time window that generate similar displacements have similar inputs to the network.
\subsection{Self-Supervised Loss}
\label{sec:loss}
Due to the fact that there is a relatively small amount of labeled data for event based cameras as compared to traditional cameras, it is difficult to generate a sufficient dataset for a supervised learning method. Instead, we utilize the fact that the DAVIS camera generates synchronized events and grayscale images to perform self-supervised learning using the grayscale images in the loss. At training time, the network is provided with the event timestamp images, as well as a pair of grayscale images, occurring immediately before and after the event time window. Only the event timestamp images are passed into the network, which predicts a per pixel flow. The grayscale images are then used to apply a loss over the predicted flow in a self-supervised manner.

The overall loss function used follows traditional variational methods for estimating optical flow, and consists of a photometric and a smoothness loss. 

To compute the photometric loss, the flow is used to warp the second image to the first image using bilinear sampling, as described in \citet{jason2016back}. The photometric loss, then, aims to minimize the difference in intensity between the warped second image and the first image:
\begin{align}
\ell_{\text{photometric}}(u,v;I_t, I_{t+1})&=\notag\\
\sum_{x, y} \rho(I_t(x, y)& - I_{t+1}(x+u(x,y), y+v(x,y)))\label{eq:photometric}
\intertext{where $\rho$ is the Charbonnier loss function, a common loss in the optical flow literature used for outlier rejection (\citet{sun2014quantitative}):}
\rho(x)=&(x^2+\epsilon^2)^{\alpha}\label{eq:charb}
\end{align}

As we are using frame based images for supervision, this method is susceptible to image-based issues such as the aperture problem. Thus, we follow the other works in the frame based domain, and apply a regularizer in the form of a smoothness loss. The smoothness loss aims to regularize the output flow by minimizing the difference in flow between neighboring pixels horizontally, vertically and diagonally.
\begin{align}
&\ell_{\text{smoothness}}(u, v)=\notag\\
&\sum_{x,y}\sum_{i,j\in \mathcal{N}(x,y)}\rho(u(x,y)-u(i, j)) + \rho(v(x,y)-v(i, j))
\end{align}
where $\mathcal{N}$ is the set of neighbors around $(x, y)$.

The total loss is the weighted sum of the photometric and smoothness losses:
\begin{align}
L_{\text{total}}=&\ell_{\text{photometric}}+\lambda \ell_{\text{smoothness}}\label{eq:total_loss}
\end{align}
\subsection{Network Architecture}
\label{sec:architecture}
The EV-FlowNet architecture very closely resembles the encoder-decoder networks such as the stacked hourglass (\citet{newell2016stacked}) and the U-Net (\citet{ronneberger2015u}), and is illustrated in Fig. \ref{fig:architecture}. The input event image is passed through 4 strided convolution layers, with output channels doubling each time. The resulting activations are passed through 2 residual blocks, and then four upsample convolution layers, where the activations are upsampled using nearest neighbor resampling and then convolved, to obtain a final flow estimate. At each upsample convolution layer, there is also a skip connection from the corresponding strided convolution layer, as well as another convolution layer to produce an intermediate, lower resolution, flow estimate, which is concatenated with the activations from the upsample convolution. The loss in \eqref{eq:total_loss} is then applied to each intermediate flow by downsampling the grayscale images. The tanh function is used as the activation function for all of the flow predictions.
\section{Optical Flow Dataset}
For ground truth evaluation only, we generated a novel dataset for ground truth optical flow using the data provided in the Multi-Vehicle Stereo Event Camera dataset (MVSEC) by \citet{zhu2018mvsec}. The dataset contains stereo event camera data in a number of flying, driving and handheld scenes. In addition, the dataset provides ground truth poses and depths maps for each event camera, which we have used to generate reference ground truth optical flow. 

From the pose (consisting of rotation $R$ and translation $\mathbf{p}$) of the camera at time $t_0$ and $t_1$, we make a linear velocity assumption, and estimate velocity and angular velocity using numerical differentiation:
\begin{align}
\mathbf{v}=&\frac{(\mathbf{p}(t_1) - \mathbf{p}(t_0))}{dt}\\
\mathbf{\omega}^{\wedge}=&\frac{\text{logm}\left(R_{t_0}^{T}R_{t_1}\right)}{dt}
\intertext{where logm is the matrix logarithm, and $\mathbf{\omega}^{\wedge}$ converts the vector $\mathbf{\omega}$ into the corresponding skew symmetric matrix:}
\mathbf{\omega}^{\wedge}=&\begin{bmatrix}0 & -\omega_z & \omega_y\\\omega_z & 0 & -\omega_x\\-\omega_y & \omega_x & 0\end{bmatrix}
\end{align}
A central moving average filter is applied to the estimated velocities to reduce noise. We then use these velocities to estimate the motion field, given the ground truth depths, $Z$, at each undistorted pixel position:
\begin{align}
\begin{pmatrix}\dot{x}\\\dot{y}\end{pmatrix}=&\begin{bmatrix}-\frac{1}{Z} & 0 & -\frac{x}{Z} & xy & -(1+x^2) & y\\0 & -\frac{1}{Z} & \frac{y}{Z} & 1+y^2 & -xy & -x\end{bmatrix}\begin{pmatrix}\mathbf{v} \\ \mathbf{\omega}\end{pmatrix}
\end{align}

Finally, we scale the motion field by the time window between each pair of images $dt$, and use the resulting displacement as an approximation to the true optical flow for each pixel. To apply the ground truth to the distorted images, we shift the undistorted pixels by the flow, and apply distortion to the shifted pixels. The distorted flow is, then, the displacement from the original distorted position to the shifted distorted position.

In total, we have generated ground truth optical flow for the indoor$\_$flying, outdoor$\_$day and outdoor$\_$night sequences. In addition to using the indoor$\_$flying and outdoor$\_$day ground truth sets for evaluation, we will also release all sequences as a dataset.
\section{Empirical Evaluation}
\label{sec:results}
\begin{figure*}[t!]
\centering
\setlength\tabcolsep{1.5pt} 
\begin{tabular}{ccccc}
Grayscale Image & Event Timestamps & Ground Truth Flow  &  UnFlow Flow &   EV-FlowNet$_{\text{2R}}$ Flow\\
  \includegraphics[width=0.20\linewidth]{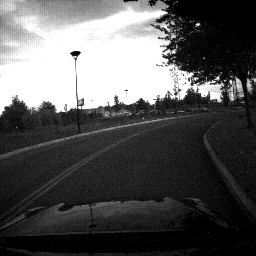} &
  \includegraphics[width=0.20\linewidth]{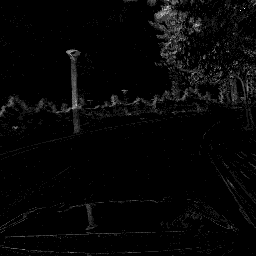} &
  \includegraphics[width=0.20\linewidth]{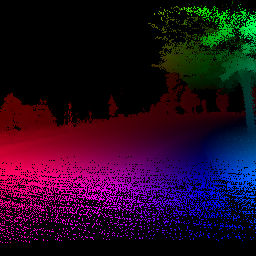} &
  \includegraphics[width=0.20\linewidth]{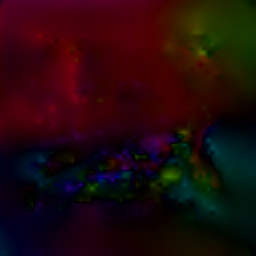} &
  \includegraphics[width=0.20\linewidth]{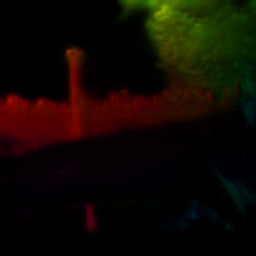}
  \\
  \includegraphics[width=0.20\linewidth]
{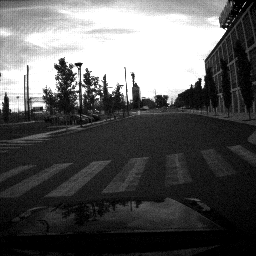} &
\includegraphics[width=0.20\linewidth]
{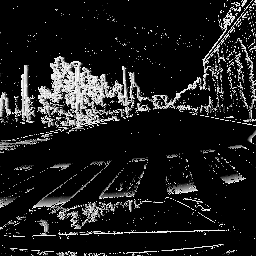} &
  \includegraphics[width=0.20\linewidth]{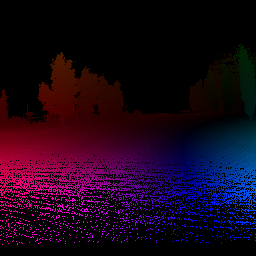} &
  \includegraphics[width=0.20\linewidth]{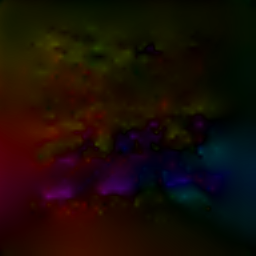} &
  \includegraphics[width=0.20\linewidth]{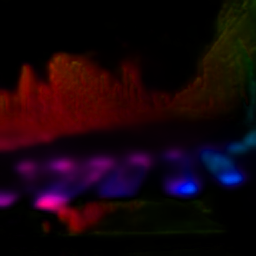}
  \\
    \includegraphics[width=0.20\linewidth]{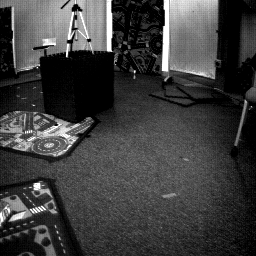} &
    \includegraphics[width=0.20\linewidth]{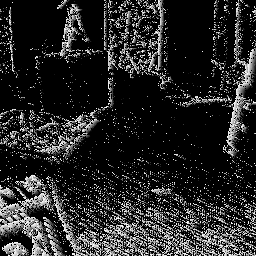} &
  \includegraphics[width=0.20\linewidth]{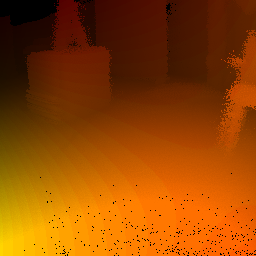} &
  \includegraphics[width=0.20\linewidth]{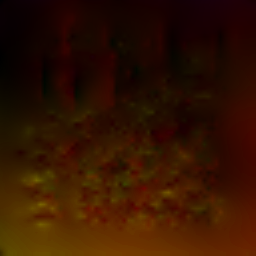} &
  \includegraphics[width=0.20\linewidth]{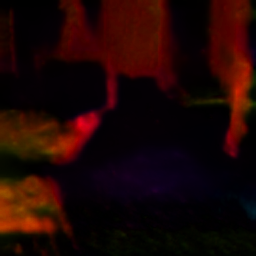}
  \\
  \includegraphics[width=0.20\linewidth]
{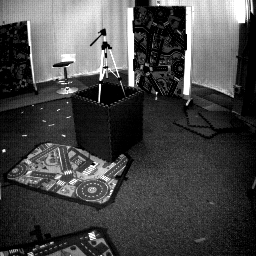} &
\includegraphics[width=0.20\linewidth]
{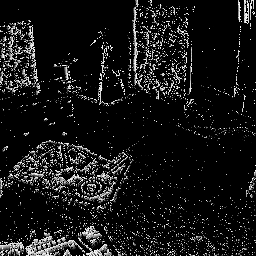} &
  \includegraphics[width=0.20\linewidth]{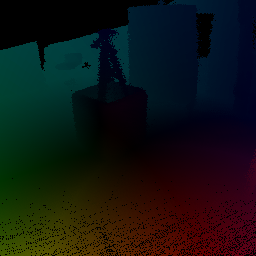} &
  \includegraphics[width=0.20\linewidth]{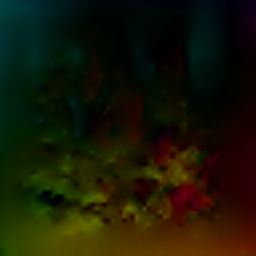} &
  \includegraphics[width=0.20\linewidth]{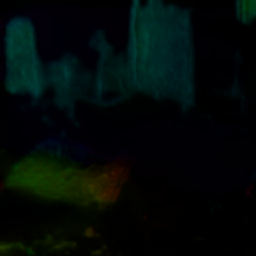}
  \end{tabular}
  \centering
  \caption{Qualitative results from evaluation. Examples were collected from outdoor$\_$day1, outdoor$\_$day1, indoor$\_$flying1 and indoor$\_$flying2, in that order. Best viewed in color.}
  \label{fig:flowfigs}
\end{figure*}
\subsection{Training Details}
Two networks were trained on the two outdoor$\_$day sequences from MVSEC. outdoor$\_$day1 contains roughly 12000 images, and outdoor$\_$day2 contains roughly 26000 images. The images are captured from driving in an industrial complex and public roads, respectively, where the two scenes are visually very different. The motions include mostly straights and turns, with occasional independently moving objects such as other cars and pedestrians. The input images are cropped to 256x256, the number of output channels at the first encoder layer is 64 and the number of output channels in each residual block is 512. 

To increase the variation in the magnitude of the optical flow seen at training, we randomly select images up to $k$ images apart in time, and all of the events that occurred between those images. In our experiments, $k\in [2, 4, 6, 8, 10, 12]$. In addition, we randomly flip the inputs horizontally, and randomly crop them to achieve the desired resolution.

The weight on the smoothness loss \eqref{eq:total_loss}, $\lambda$, is set to 0.5. Each of the intermediate losses is weighted equally in the final loss. For the Charbonnier loss \eqref{eq:charb}, $\alpha$ was set to be 0.45 and $\epsilon$ was set to be 1e-3. The Adam optimizer is used, with learning rate initialized at 1e-5, and exponentially decayed every 4 epochs by 0.8. The model is trained for 300,000 iterations, and takes around 12 hours to train on a 16GB NVIDIA Tesla V100.
\subsection{Ablation Studies}
In addition to the described architecture (denoted EV-FlowNet$_{\text{2R}}$), we also train three other networks to test the effects of varying the input to the network, as well as increasing the capacity of the network.

To test the contribution of each of the channels in the input, we train two additional networks, one with only the event counts (first two channels) as input (denoted EV-FlowNet$_{\text{C}}$), and one with only the event timestamps (last two channels) as input (denoted EV-FlowNet$_{\text{R}}$).

In addition, we tested different network capacities by training a larger model with 4 residual blocks (denoted EV-FlowNet$_{\text{4R}}$). A single forward pass takes, on average, 40ms for the smaller network, and 48ms for the larger network, when run on a NVIDIA GeForce GTX 1050, a laptop grade GPU.

\begin{table*}[t!]
\centering
\begin{tabular}{cccccccccccc} 
\hline
dt=1 frame & \multicolumn{2}{c}{outdoor driving} & & \multicolumn{2}{c}{indoor flying1} & & \multicolumn{2}{c}{indoor flying2} & & \multicolumn{2}{c}{indoor flying3}\\
 
\cline{2-3} \cline{5-6} \cline{8-9} \cline{11-12}

& AEE & $\%$ Outlier & & AEE & $\%$ Outlier & & AEE & $\%$ Outlier & & AEE & $\%$ Outlier\\

\hline
UnFlow
& 0.97  & 1.6  & & \textbf{0.50}  &  \textbf{0.1} &  &  \textbf{0.70} & \textbf{1.0} & & \textbf{0.55}& \textbf{0.0}\\
EV-FlowNet$_{\text{C}}$
& \textbf{0.49}  & \textbf{0.2}  &   & 1.30  &  6.8 &  & 2.34 & 25.9 & & 2.06 & 22.2\\
EV-FlowNet$_{\text{T}}$
& 0.52 & \textbf{0.2}  &   &  1.20 & 4.5 &   & 2.15  & 22.6  & & 1.91 & 19.8\\
EV-FlowNet$_{\text{2R}}$
&  \textbf{0.49} & \textbf{0.2} & & 1.03  & 2.2 & & 1.72 & 15.1  & & 1.53 & 11.9\\
EV-FlowNet$_{\text{4R}}$
& \textbf{0.49}  & \textbf{0.2}  &   &  1.14 & 3.5 &   & 2.10  & 21.0  & & 1.85 & 18.8\\
\hline
dt=4 frames& \multicolumn{2}{c}{outdoor driving} & & \multicolumn{2}{c}{indoor flying1} & & \multicolumn{2}{c}{indoor flying2} & & \multicolumn{2}{c}{indoor flying3}\\
 
\cline{2-3} \cline{5-6} \cline{8-9} \cline{11-12}

& AEE & \% Outlier & & AEE & \% Outlier & & AEE & \% Outlier & & AEE & \% Outlier\\

\hline
UnFlow
& 2.95  & 40.0 & & 3.81 & 56.1 & & 6.22 & 79.5 &  & \textbf{1.96} & \textbf{18.2} \\
EV-FlowNet$_{\text{C}}$
&  1.41 & 10.8  &   &  3.22 & 41.4 &   &  5.30 & 60.1  & & 4.68 & 57.0\\
EV-FlowNet$_{\text{T}}$
& 1.34  & 8.4  &   &  2.53 & 33.7  &   & 4.40  & 51.9  & & 3.91 & 47.1\\
EV-FlowNet$_{\text{2R}}$
& \textbf{1.23}  &  \textbf{7.3} &  & \textbf{2.25}  & \textbf{24.7} & & \textbf{4.05}  & \textbf{45.3} & & 3.45 & 39.7\\
EV-FlowNet$_{\text{4R}}$
& 1.33  & 9.4  &   &  2.75 & 33.5 &   & 4.82  &  53.3 & & 4.30 & 47.8\\
\end{tabular}
\caption{Quantitative evaluation of each model on the MVSEC optical flow ground truth. Average end-point error (AEE) and percentage of pixels with EE above 3 and 5\% of the magnitude of the flow vector(\% Outlier) are presented for each method (lower is better for both), with evaluation run with image pairs 1 frame apart (top) and 4 frames apart (bottom). The EV-FlowNet methods are: Counts only (EV-FlowNet$_{\text{c}}$), Timestamps only (EV-FlowNet$_{\text{T}}$), 2 Residual blocks (EV-FlowNet$_{\text{2R}})$ and 4 Residual blocks (EV-FlowNet$_{\text{4R}}$).}
\label{tab:results}
\end{table*}
\subsection{Comparisons}
To compare our results with other existing methods, we tested implementations of Event-based Visual Flow by \citet{benosman2014event}, an optimization based method that works on events, and UnFlow by \citet{meister2017unflow}, a self supervised method that works on traditional frames.

As there is no open source code by the authors of Event-based Visual Flow, we designed an implementation around the method described in \citet{rueckauer2016evaluation}. In particular, we implemented the robust Local Plane Fit algorithm, with a spatial window of 5x5 pixels, vanishing gradient threshold th3 of 1e-3, and outlier distance threshold of 1e-2. However, we were unable to achieve any reasonable results on the datasets, with only very few points returning valid flow values ($<5\%$), and none of the valid flow values being visually correct. For validation, we also tested the open source MATLAB code provided by the authors of \citet{mueggler2017event}, where we received similar results. As a result, we believe that the method was unable to generalize to the natural scenes in the test set, and so did not include the results in this paper.

For UnFlow, we used the unsupervised model trained on KITTI raw, and fine tuned on outdoor$\_$day2. This model was able to produce reasonable results on the testing sets, and we include the results in the quantitative evaluation in Tab. \ref{tab:results}.

\subsection{Test Sequences}
For comparison against UnFlow, we evaluated 800 frames from the outdoor$\_$day1 sequence as well as sequences 1 to 3 from indoor$\_$flying. For the event input, we used all of the events that occurred in between the two input frames.

The outdoor$\_$day1 sequence spans between 222.4s and 240.4s. This section was chosen as the grayscale images were consistently bright, and there is minimal shaking of the camera (the provided poses are smoothed and do not capture shaking of the camera if the vehicle hits a bump in the road). In order to avoid conflicts between training and testing data, a model trained only using data from outdoor$\_$day2 was used, which is visually significantly different from outdoor$\_$day1. 

The three indoor$\_$flying sequences total roughly 240s, and feature a significantly different indoor scene, containing vertical and backward motions, which were previously unseen in the driving scenes. A model trained on both outdoor$\_$day1 and outdoor$\_$day2 data was used for evaluation on these sequences. We avoided fine tuning on the flying sequences, as the sequences are in one room, and all relatively similar in visual appearance. As a result, it would be very easy for a network to overfit the environment. Sequence 4 was omitted as the majority of the view was just the floor, and so had a relatively small amount of useful data for evaluation.

\subsection{Metrics}
For each method and sequence, we compute the average endpoint error (AEE), defined as as the distance between the endpoints of the predicted and ground truth flow vectors:
\begin{align}
AEE=&\sum_{x,y}\left\|\begin{pmatrix}u(x,y)_{\text{pred}} \\ v(x,y)_{\text{pred}}\end{pmatrix} - \begin{pmatrix}u(x,y)_{\text{gt}} \\ v(x,y)_{\text{gt}}\end{pmatrix}\right\|_2
\end{align}
In addition, we follow the KITTI flow 2015 benchmark and report the percentage of points with EE greater than 3 pixels and 5\% of the magnitude of the flow vector. Similarly to KITTI, 3 pixels is roughly the maximum error observed when warping the grayscale images according to the ground truth flow, and comparing against the next image.

However, as the input event image is relatively sparse, the network only returns accurate flow on points with events. As a result, we limit the computation of AEE to pixels in which at least one event was observed. For consistency, this is done with a mask applied to the EE for both event-based and frame-based methods. We also mask out any points for which we have no ground truth flow (i.e. regions with no ground truth depth). In practice, this results in the error being computed over 20-30\% of the pixels in each image. 

In order to vary the magnitude of flow observed for each test, we run two evaluations per sequence: one with input frames and corresponding events that are one frame apart, and one with frames and events four frames apart. We outline the results in Tab. \ref{tab:results}.
\subsection{Results}
\begin{figure}[t!]
\centering
\includegraphics[width=0.32\linewidth]{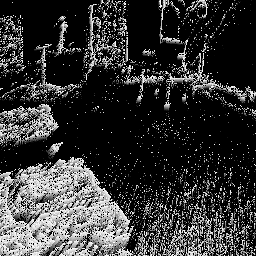}
  \includegraphics[width=0.32\linewidth]{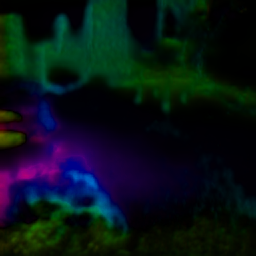}
  \includegraphics[width=0.32\linewidth]{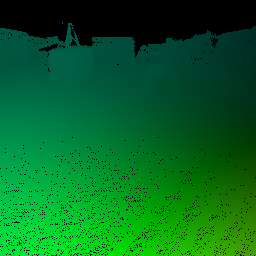}
  \centering
  \caption{Common failure case, where fast motion causes recent timestamps to overwrite older pixels nearby, resulting in incorrect predictions. Best viewed in color.}
  \label{fig:failurecase}
\end{figure}
\subsubsection{Qualitative Results}
In addition to the quantitative analysis provided, we provide qualitative results in Fig. \ref{fig:flowfigs}. In these results, and throughout the test set, the predicted flow always closely follows the ground truth. As the event input is quite sparse, our network tends to predict zero flow in areas without events. This is consistent with the photometric loss, as areas without events are typically low texture areas, where there is little change in intensity within each pixel neighborhood. In practice, the useful flow can be extracted by only using flow predictions at points with events. On the other hand, while UnFlow typically performs reasonably on the high texture regions, the results on low texture regions are very noisy. 
\subsubsection{Ablation Study Results}
From the results of the ablation studies in Tab. \ref{tab:results}, EV-FlowNet$_{\text{C}}$ (counts only) performed the worst. This aligns with our intuition, as the only information attainable from the counts is from motion blur effects, which is a weak signal on its own. EV-FlowNet$_{\text{T}}$ (timestamps only) performs better for most tests, as the timestamps carry information about the ordering between neighboring events, as well as the magnitude of the velocity. However, the timestamp only network fails when there is significant noise in the image, or when fast motion results in more recent timestamps covering all of the older ones. This is illustrated in Fig \ref{fig:failurecase}, where even the full network struggles to predict the flow in a region dominated by recent timestamps. Overall, the combined models clearly perform better, likely as the event counts carry information about the importance of each pixel. Pixels with few events are likely to be just noise, while pixels with many events are more likely to carry useful information. 
Somewhat surprisingly, the larger network, EV-FlowNet$_{\text{4R}}$ actually performs worse than the smaller one, EV-FlowNet$_{\text{2R}}$. A possible explanation is that the larger capacity network learned to overfit the training sets, and so did not generalize as well to the test sets, which were significantly different. For extra validation, both EV-FlowNet$_{\text{2R}}$ and EV-FlowNet$_{\text{4R}}$ were trained for an additional 200,000 iterations, with no appreciable improvements. It is likely, however, that, given more data, the larger model would perform better. 
\subsubsection{Comparison Results}
From our experiments, we found that the UnFlow network tends to predict roughly correct flows for most inputs, but tends to be very noisy in low texture areas of the image. The sparse nature of the events is a benefit in these regions, as the lack of events there would cause the network to predict no flow, instead of an incorrect output.

In general, EV-FlowNet performed better on the dt=4 tests, while worse on the dt=1 tests (with the exception of outdoor$\_$driving1 and indoor$\_$flying3). We observed that UnFlow typically performed better in situations with very small or very large motion. In these situations, there are either few events as input, or so many events that the image is overriden by recent timestamps. However, this is a problem intrinsic to the testing process, as the time window is defined by the image frame rate. In practice, these problems can be avoided by choosing time windows large enough so that sufficient information is available while avoiding saturating the event image. One possible solution to this would be to have a fixed number of events in the window each time. 
\section{Conclusion}
\label{sec:conclusion}
In this work, we have presented a novel design for a neural network architecture that is able to accurately predict optical flow from events alone. Due to the method's self-supervised nature, the network can be trained without any manual labeling, simply by recording data from the camera. We show that the predictions generalize beyond hand designed laboratory scenes to natural ones, and that the method is competitive with state of the art frame-based self supervised methods. We hope that this work will provide not only a novel method for flow estimation, but also a paradigm for applying other self-supervised learning methods to event cameras in the future. For future work, we hope to incorporate additional losses that provide supervisory signals from event data alone, to expose the network to scenes that are challenging for traditional frame-based cameras, such as those with high speed motions or challenging lighting.

\section*{Acknowledgments}
Thanks to Tobi Delbruck and the team at iniLabs for providing and supporting the DAVIS-346b cameras. We also gratefully appreciate support through the following grants: NSF-IIS-1703319, NSF-IIP-1439681 (I/UCRC),  ARL RCTA W911NF-10-2-0016, and the DARPA FLA program.
\addtolength{\textheight}{-400pt}
\bibliographystyle{plainnat}
\bibliography{references}

\end{document}